\documentclass[10pt,journal,letterpaper,compsoc]{IEEEtran}
\usepackage[cmex10]{amsmath}
\usepackage{amsfonts}
\usepackage{amssymb}
\usepackage{bm}
\usepackage{graphicx}
\usepackage{graphics}
\usepackage[applemac]{inputenc}
\usepackage{cite}
\usepackage{mdwtab}
\usepackage{subfigure}
\usepackage{color}
\usepackage{xspace}
\usepackage{url}
\usepackage{tabularx}
\usepackage{epstopdf}
\usepackage{multirow}

\usepackage{caption}

\newcolumntype{C}[1]{>{\centering}m{#1}}

\makeatletter
\DeclareRobustCommand\onedot{\futurelet\@let@token\@onedot}
\def\@onedot{\ifx\@let@token.\else.\null\fi\xspace}

\def\eg{\emph{e.g}\onedot} 
\def\ie{\emph{i.e}\onedot} 
\def\cf{\emph{cf}\onedot}

\def\etal{\emph{et al}\onedot}
\makeatother

\newcommand{\Fig}{Fig.\xspace}

\newcommand{\Sec}{Sec.\xspace}
\newcommand{\Tab}{Tab.\xspace}

\newcolumntype{L}[1]{>{\raggedright\let\newline\\\arraybackslash\hspace{0pt}}m{#1}}
\newcolumntype{C}[1]{>{\centering\let\newline\\\arraybackslash\hspace{0pt}}m{#1}}
\newcolumntype{R}[1]{>{\raggedleft\let\newline\\\arraybackslash\hspace{0pt}}m{#1}}

\newcommand{\MOTChallenge}{{\it MOTChallenge}\xspace}
\newcommand{\MOTNEW}{{\it MOT16}\xspace}
\newcommand{\MOTOLD}{{\it MOT15}\xspace}

\newcommand{\dismeas}{d}			
\newcommand{\simthresh}{t_d}  		

\newcommand{\parvec}{\Theta}		
\newcommand{\onepar}{\theta}		

\newcommand{\Tracker}[1]{\textsc{\lowercase{#1}}\xspace}

\graphicspath{{figures/},{figures/moving/},{figures/static/},{figures/annotations/}}

\definecolor{darkgreen}{rgb}{0,.75,0}
\definecolor{gray40}{gray}{.40}

   \teaser{
   \centering
   \hspace{0.005\linewidth}
   \includegraphics[width=.92\linewidth]{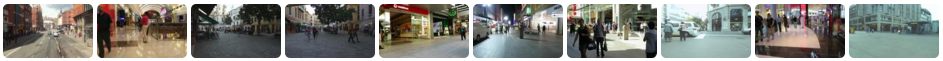}
  }

\begin{document}
\title{MOT16: A Benchmark for Multi-Object Tracking}

\author{        Anton Milan$^*$,
    Laura Leal-Taix{\'e}$^*$,
        Ian Reid, 
        Stefan Roth,
        and Konrad Schindler
\thanks{$^*$ = authors contributed equally.}
\thanks{A.~Milan$^*$ and I.~Reid are with the Australian Centre for Visual Technologies at the University of Adelaide, Australia.}
\thanks{L.~Leal-Taix{\'e}$^*$ and K.~Schindler are with the Photogrammetry and Remote Sensing Group at ETH Zurich, Switzerland.}
\thanks{S.~Roth is with the Department of Computer Science, Technische Universit{\"a}t Darmstadt, Germany.}
\thanks{Primary contacts: leal@geod.baug.ethz.ch, anton.milan@adelaide.edu.au }
}

\IEEEcompsoctitleabstractindextext{%

\begin{abstract}

Standardized benchmarks are crucial for the majority of computer vision applications.
Although leaderboards and ranking tables should not be over-claimed, benchmarks often
provide the most objective measure of performance and are therefore important guides 
for reseach.
Recently, a new benchmark for Multiple Object Tracking, \MOTChallenge, was 
launched with the goal of collecting existing and new data and creating a framework
for the standardized evaluation of multiple object tracking methods \cite{MOTChallenge:arxiv:2015}. 
The first release of the benchmark focuses on multiple people tracking, since
pedestrians are by far the most studied object in the tracking community. 
This paper accompanies a new release of the \MOTChallenge benchmark.
Unlike the initial release, all videos of \MOTNEW have been carefully
annotated following a consistent protocol. Moreover, it not only offers a significant increase
in the number of labeled boxes, but also provides multiple object classes beside pedestrians
and the level of visibility for every single object of interest.

\end{abstract}

}

\maketitle

\IEEEdisplaynotcompsoctitleabstractindextext

\IEEEpeerreviewmaketitle


\section{Introduction}
\label{sec:introduction}

Evaluating and comparing multi-target tracking methods is not trivial 
for numerous reasons (\emph{cf.~e.g.}~\cite{Milan:2013:CVPRWS}). 
First, unlike for other tasks, such as image denoising, the ground 
truth, \emph{i.e.}~the perfect solution one aims to achieve, is
difficult to define clearly. Partially visible, occluded, or cropped targets, 
reflections in mirrors or windows, and objects that very closely resemble 
targets all impose intrinsic ambiguities, such that even humans may 
not agree on one particular ideal solution. Second, a number of different
evaluation metrics with free parameters and ambiguous definitions often 
lead to conflicting quantitative results across the literature. Finally, the 
lack of pre-defined test and training data makes it difficult to compare
different methods fairly.

Even though multi-target tracking is a crucial problem in scene understanding, 
until recently it still lacked large-scale benchmarks to provide a fair comparison between 
tracking methods. 
In 2014, we released the \MOTChallenge benchmark, which consisted  of
three main components:
(1) a (re-)collection of publicly available and new datasets, (2) a centralized evaluation method, and (3) 
an infrastructure that allows for crowdsourcing of new data, new 
evaluation methods and even new annotations. 
The first release of the dataset named \MOTOLD consisted of 11 sequences for training and 11 for testing, with a total 
of 11286 frames or 996 seconds of video.  
Pre-computed object detections, annotations (only for the training sequences), and a common evaluation method for 
all datasets was provided to all participants, which allowed for all results to be compared in a fair way. 

%

Since October 2014, 47 methods have been publicly tested on the \MOTChallenge benchmark, and over 180 users have registered. It has been established as a new standard benchmark for multiple people tracking, and methods have improved accuracy by over 10\%. The first workshop \cite{bmtt2015} organized on the \MOTChallenge benchmark took place in 
early 2015 in conjunction with the Winter Conference on Applications of 
Computer Vision (WACV).
Despite its success, \MOTOLD is lacking in a few aspects:
\begin{itemize}
\item The annotation protocol is not consistent across all sequences since some of the ground truth was collected from various sources with already available annotations;
\item the distribution of crowd density is not balanced for training and test sequences;
\item some of the sequences are easy and well-known (\eg PETS09-S2L1) and methods are overfitted to them, which makes them not ideal for training
 purposes;
 \item the provided detections did not show good performance on the benchmark, which made some participants switch to another pedestrian detector.
\end{itemize}  

In order to improve the above shortcomings, we now introduce the new \MOTNEW benchmark, a set of 14 sequences with more crowded scenarios, different viewpoints, camera motions and weather conditions. Most importantly, the annotations for \emph{all} sequences have been carried out by qualified researchers from scratch following a strict protocol, and finally double-checked to ensure highest annotation accuracy. Not only pedestrians are annotated, but also vehicles, sitting people, occluding objects, as well as other significant object classes. With this fine-grained level of annotation it is possible to accurately compute the degree of occlusion and cropping of all bounding boxes, which is also provided with the benchmark. We hope that this rich ground truth information will be very useful to the community in order to develop even more accurate tracking methods and advancing the field further. 
%
\noindent
This paper has thus three main goals:

\begin{enumerate}
 \item To present the new \MOTNEW benchmark for fair evaluation of multi-target tracking methods;
	\item to detail the annotation protocol strictly followed to create the ground truth of the benchmark;
  \item to bring forward the strengths and weaknesses of state-of-the-art 
multi-target tracking methods.
 
\end{enumerate}

The benchmark with all datasets, current ranking and submission 
guidelines can be found at:
\begin{center}
\url{http://www.motchallenge.net/}
\end{center}

\subsection{Related work}
\label{sec:related-work}
\noindent{\bf Benchmarks and challenges.}
In the recent past, the computer vision community has developed
centralized benchmarks for numerous tasks including object detection 
\cite{Everingham:2012:VOC}, pedestrian detection 
\cite{caltechpedestrians}, 3D reconstruction \cite{Seitz:2006:CVPR}, 
optical flow \cite{Baker:2011:IJCV,Geiger:2012:CVPR}, visual odometry 
\cite{Geiger:2012:CVPR}, single-object short-term tracking 
\cite{VOC2014}, and stereo estimation \cite{Scharstein:2002:IJCV, 
Geiger:2012:CVPR}. Despite potential pitfalls of such benchmarks (\eg~\cite{Torralba:2011:ULD}), they 
have proven to be extremely helpful to advance the state of the art in 
the respective area. For multiple target tracking, in contrast, there
has been very limited work on standardizing quantitative
evaluation.

One of the few exceptions is the well known PETS dataset 
\cite{Ferryman:2010:PETS}, targeted primarily at surveillance 
applications. The 2009 version consisted of 3 subsets: S1 targeted 
at person count and density estimation, S2 targeted at people 
tracking, and S3 targeted at flow analysis and event recognition. The 
easiest sequence for tracking (S2L1) consisted of a scene with few pedestrians, 
and for that sequence state-of-the-art methods perform extremely well 
with accuracies of over 90\% given a good set of initial detections 
\cite{Milan:2014:PAMI, Henriques:2011:ICCV, Zamir:2012:ECCV}. Methods then 
moved to tracking on the hardest sequence (\ie with the highest crowd
density), but hardly ever on the complete dataset. Even for this
widely used benchmark, we observe that tracking results are commonly
obtained in an inconsistent fashion: involving using different subsets of 
the available data, inconsistent model training that is often prone to overfitting, varying evaluation scripts, and different detection inputs.
Results are thus not easily comparable.  
Hence, the question that arises is: Are these sequences already too easy 
for current tracking methods, are methods simply overfit, or are
they poorly evaluated?

The PETS team organizes a workshop approximately once a year to which
researchers can submit their results, and methods are evaluated
under the same conditions. Although this is indeed a fair comparison,
the fact that submissions are evaluated only once a year means that
the use of this benchmark for high impact conferences like ICCV or
CVPR remains challenging.
Furthermore, the sequences tend to be focused only on surveillance scenarios, and
lately on specific tasks such as vessel tracking.

A well-established and useful way of organizing datasets is through 
standardized challenges. These are usually in the form of web servers 
that host the data and through which results are uploaded by the
users.
Results are then computed in a centralized way by the server and
afterwards presented online to the public, making comparison with any
other method immediately possible.
There are several datasets organized in this fashion: the Labeled Faces 
in the Wild \cite{huangtech2007} for unconstrained face recognition, the 
PASCAL VOC \cite{Everingham:2012:VOC} for object detection, the ImageNet 
large scale visual recognition challenge \cite{Russakovsky:2014:arxiv},
or the Reconstruction Meets Recognition Challenge (RMRC) \cite{rmrc2014}.


Recently, the KITTI benchmark \cite{Geiger:2012:CVPR} was introduced for 
challenges in autonomous driving, which included stereo/flow, odometry, 
road and lane estimation, object detection and orientation estimation, as 
well as tracking. Some of the sequences include crowded pedestrian 
crossings, making the dataset quite challenging, but the camera position 
is always the same for all sequences (at a car's height).



Another work that is worth mentioning is \cite{alahicvpr2014}, in which 
the authors collected a very large amount of data with 42 million pedestrian 
trajectories. Since annotation of such a large collection of data is 
infeasible, they use a denser set of cameras to create the ``ground 
truth'' trajectories. Though we do not aim at collecting such a large 
amount of data, the goal of our benchmark is somewhat similar: to push research 
in tracking forward by generalizing the test data to a larger set that is 
highly variable and hard to overfit. 


In the near future, DETRAC, a new benchmark for vehicle tracking \cite{detracarxiv2015}, is going to 
open a similar submission system to the one we proposed with \MOTChallenge. The benchmark consists of a total of 
100 sequences, 60\% of which is used for training. Sequences are filmed from a high viewpoint (surveillance scenarios)
with the goal of vehicle tracking.

With the \MOTNEW release within the \MOTChallenge benchmark, we aim to increase the difficulty 
by including a variety of sequences filmed from different 
viewpoints, with different lighting conditions, and far more crowded scenarios when compared to our first release.

\newcommand{\thumbwidth}{0.135\linewidth}
\newcommand{\thumbheight}{0.08\linewidth}
\begin{figure*}
\centering
 \includegraphics[width=\thumbwidth,height=\thumbheight]{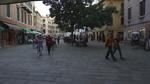}
 \includegraphics[width=\thumbwidth,height=\thumbheight]{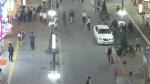}
 \includegraphics[width=\thumbwidth,height=\thumbheight]{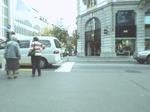}
 \includegraphics[width=\thumbwidth,height=\thumbheight]{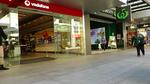}
 \includegraphics[width=\thumbwidth,height=\thumbheight]{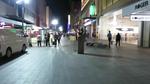}
 \includegraphics[width=\thumbwidth,height=\thumbheight]{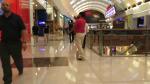}
 \includegraphics[width=\thumbwidth,height=\thumbheight]{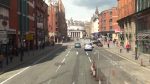}\\[1.em]
  \includegraphics[width=\thumbwidth,height=\thumbheight]{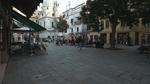}
  \includegraphics[width=\thumbwidth,height=\thumbheight]{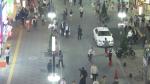}
  \includegraphics[width=\thumbwidth,height=\thumbheight]{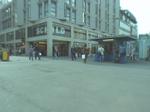}
  \includegraphics[width=\thumbwidth,height=\thumbheight]{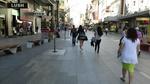}
  \includegraphics[width=\thumbwidth,height=\thumbheight]{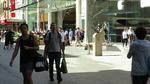}
  \includegraphics[width=\thumbwidth,height=\thumbheight]{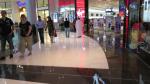}
  \includegraphics[width=\thumbwidth,height=\thumbheight]{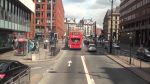}
  \caption{An overview of the \MOTNEW dataset. Top: Training sequences; bottom: test sequences.}
 \label{fig:thumbnails}
\end{figure*}

\bigskip


\noindent{\bf Evaluation.} 
A critical point with any dataset is how to measure the performance of 
the algorithms. In the case of multiple object tracking, the CLEAR 
metrics \cite{clear} have emerged as one of the standard measures.
We will discuss them in more detail in \Sec~\ref{sec:evaluation-metrics}.
By measuring the intersection over union of bounding boxes and matching those 
from ground truth annotations and results, measures of accuracy and precision can be 
computed. Precision measures how well the persons are localized, while 
accuracy evaluates how many distinct errors such as missed targets, ghost
trajectories, or identity switches are made.

Another set of measures that is widely used in the tracking community is 
that of \cite{Li:2009:CVPR}. There are three widely used metrics 
introduced in that work: mostly tracked, mostly lost, and partially
tracked pedestrians. These numbers give a very good intuition on the  
performance of the method. We refer the reader to 
\cite{MOTChallenge:arxiv:2015}
for more formal definitions.

A key parameter in both families of metrics is the intersection-over-union 
threshold, which determines if a bounding box is matched to an 
annotation or not. It is fairly common to observe methods compared under 
different thresholds, varying from 25\% to 50\%. There are often many other 
variables and implementation details that differ between 
evaluation scripts, but which may affect results significantly. 
%

It is therefore clear that standardized benchmarks are the only way to 
compare methods in a fair and principled way. Using the same ground 
truth data and evaluation methodology is the only way to 
guarantee that the only part being evaluated is the tracking method that 
delivers the results. This is the main goal behind this paper and behind 
the \MOTChallenge benchmark.

\section{Annotation rules}
\label{sec:anno-rules}

We follow a set of rules to annotate every moving person or vehicle within each sequence 
with a bounding box as accurately as possible. In the following we 
define a clear protocol that was obeyed throughout the entire dataset to 
guarantee consistency.

\subsection{Target class}
In this benchmark we are interested in tracking moving objects in videos. In particular, we are interested in evaluating multiple people tracking algorithms, therefore, people will be the center of attention of our annotations. 
We divide the pertinent classes into three categories: \\
(i) {\it moving} or {\it standing} pedestrians;\\
 (ii) people that are {\it not in an upright position} or artificial representations of humans;  and \\
 (iii) {\it vehicles} and {\it occluders}.
 
In the first group, we annotate all moving or standing (upright) pedestrians that appear in the field of view and can be determined as such by the viewer. People on bikes or skateboards will also be annotated in this category (and are typically found by modern pedestrian detectors). Furthermore, if a person \emph{briefly} bends over or squats, \eg to pick something up or to talk to a child, they shall remain in the standard \emph{pedestrian} class.
The algorithms that submit to our benchmark are expected to track these targets.

In the second group we include all people-like objects whose exact classification is ambiguous and can vary depending on the viewer, the application at hand, or other factors. We annotate all static people that are not in an upright position, \eg sitting, lying down. We also include in this category any artificial representation of a human that might fire a detection response, such as mannequins, pictures, or reflections. People behind glass should also be marked as distractors.
The idea is to use these annotations in the evaluation such that an algorithm is neither penalized nor rewarded for tracking, \eg, a sitting person or a reflection.

In the third group, we annotate all moving vehicles such as cars, bicycles, motorbikes and non-motorized vehicles (\eg strollers), as well as other potential occluders. These annotations will not play any role in the evaluation, but are provided to the users both for training purposes and for computing the level of occlusion of pedestrians. Static vehicles (parked cars, bicycles) are not annotated as long as they do not occlude any pedestrians.

%
%

\begin{table}[t]
\begin{tabular}{lp{.75\linewidth}}
& Instruction\\
\hline
What? & Targets: All upright people including\\
& + walking, standing, running pedestrians\\
& + cyclists, skaters\\ [1em]
& Distractors: Static people or representations\\
& + people not in upright position (sitting, lying down)\\
& + reflections, drawings or photographs of people\\
& + human-like objects like dolls, mannequins\\[1em]
& Others: Moving vehicles and occluders\\
& + Cars, bikes, motorbikes\\
& + Pillars, trees, buildings\\
\hline
When? & Start as early as possible.\\
& End as late as possible.\\
& Keep ID as long as the person is inside the field of view and its path can be determined unambiguously.\\
\hline
How? & The bounding box should contain all pixels belonging to that person and at the same time be as tight as possible.\\
\hline
Occlusions & Always annotate during occlusions if the position can be determined unambiguously. \\
& If the occlusion is very long and it is not possible to determine the path of the object using simple reasoning (\eg constant velocity assumption), the object will be assigned a new ID once it reappears. \\
\hline
\end{tabular}
\caption{Instructions obeyed during annotations.}
\label{tab:instructions}
\end{table}

\begin{figure*}
\centering
 \includegraphics[width=0.6\linewidth]{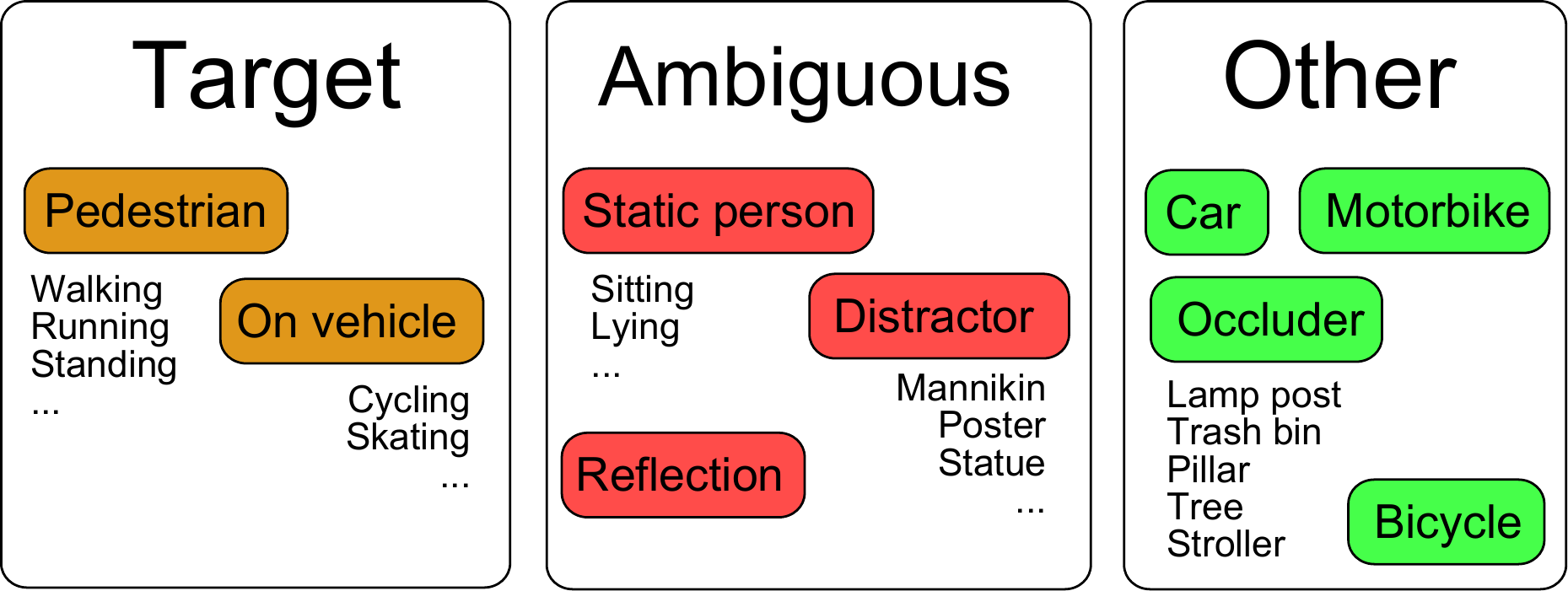}
  \includegraphics[width=0.38\linewidth]{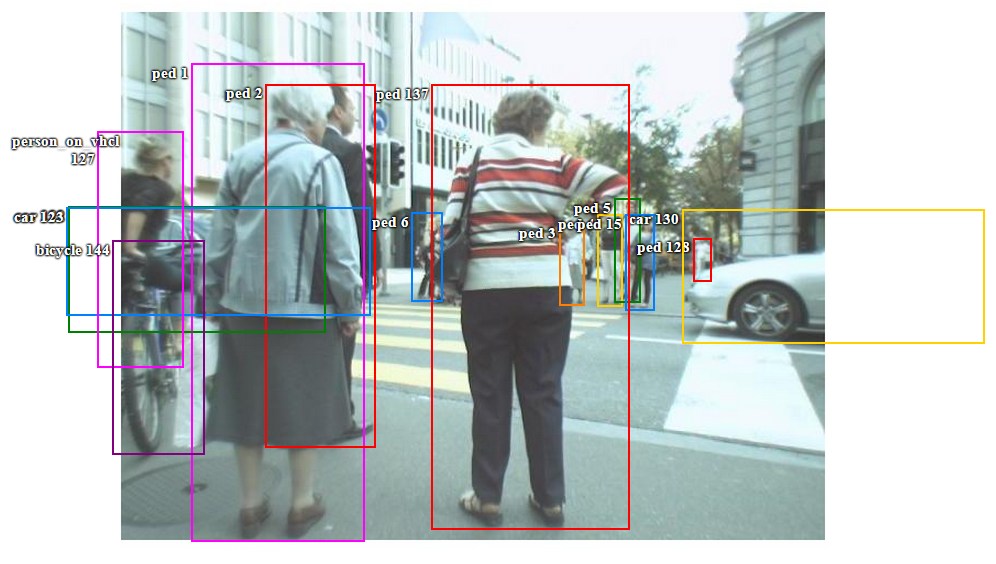}
 \caption{Left: An overview of annotated classes. The classes in orange will be the central ones to evaluate on. The red classes include ambiguous cases such that both recovering and missing them will not be penalized in the evaluation. The classes in green are annotated for training purposes and for computing the occlusion level of all pedestrians. Right: An exemplar of an annotated frame. Note how partially cropped objects are also marked outside of the frame. Also note that the bounding box encloses the entire person but not \eg the white bag of Pedestrian 1 (bottom left).}
 \label{fig:class}
\end{figure*}

The rules are summarized in \Tab~\ref{tab:instructions} and in \Fig~\ref{fig:class} we present a diagram of the classes of objects we annotate, as well as a sample frame with annotations. 

\subsection{Bounding box alignment}
The bounding box is aligned with the object's extent as accurately as possible. 
The bounding box should contain all pixels belonging to that object and at the same time be as tight as possible, \ie no pixels should be left outside the box. 
This means that a walking side-view pedestrian will typically have a box whose width varies periodically with the stride, while a front view or a standing person will maintain a more constant aspect ratio over time. If the person is partially occluded, the extent is estimated 
based on other available information such as expected size, shadows, 
reflections, previous and future frames and other cues. If a person is 
cropped by the image border, the box is estimated beyond the original frame to represent the entire person and to estimate the level of cropping. If an occluding object cannot be accurately enclosed in one box (\eg a tree with branches or an escalator may require a large bounding box where most of the area does not belong to the actual object), then several boxes may be used to better approximate the extent of that object.

Persons on vehicles will only be annotated separately from the vehicle if clearly visible. For example, children inside strollers or people inside cars will not be annotated, while motorcyclists or bikers will be.

\subsection{Start and end of trajectories}
The box (track) appears as soon as the person's location and extent can 
be determined precisely. This is typically the case when $\approx 10 \%$ of the person becomes visible.
  Similarly, the track ends when it is no longer 
possible to pinpoint the exact location. In other words the annotation 
starts as early and ends as late as possible such that the accuracy is 
not forfeited. The box coordinates may exceed the visible area.
Should a person leave the field of view and appear at a 
later point, they will be assigned a new ID.

\subsection{Minimal size}
Although the evaluation will only take into account pedestrians that have a minimum height in pixels, 
annotations will contain all objects of all sizes as long as they are distinguishable by the annotator. In other words, \emph{all} targets independent of their size on the image shall be annotated.

\subsection{Occlusions}
There is no need to explicitly annotate the level of occlusion. This value will be computed automatically using the ground plane assumption and the annotations. Each target is fully annotated through occlusions as long as its extent and location can be determined accurately enough. If a target becomes completely occluded in the middle of the sequence and does not become visible later, the track should be terminated (marked as `outside of view'). If a target reappears after a prolonged period such that its location is ambiguous during the occlusion, it will reappear with a new ID.

\subsection{Sanity check}
Upon annotating all sequences, a ``sanity check'' was carried out to ensure that no relevant entities were missed. To that end, we ran a pedestrian detector on all videos and added all high-confidence detections that corresponded to either humans or distractors to the annotation list.

\section{Datasets}
\label{sec:datasets}

One of the key aspects of any benchmark is data collection. The goal of \MOTNEW is to compile a benchmark with new sequences, which are more challenging than the ones presented in \MOTOLD. 
In \Fig~\ref{fig:thumbnails} and \Tab~\ref{tab:dataoverview} we show an overview of the sequences 
included in the benchmark.

\begin{table*}[tbh]
\begin {center}
 \begin{tabular}{|l| c| c| r| r| r| r| c |c | c | c| }
 \hline
 \multicolumn{11}{|c|}{\bf Training sequences} \\ 
 \hline 
      Name & FPS & Resolution & Length & Tracks & Boxes & Density & Camera & Viewpoint & Conditions & Source\\ 
      \hline
     MOT16-02 & 30 & 1920x1080 & 600 (00:20) & 49 & 17,833 & 29.7  & static  & medium & cloudy & new \\
     MOT16-04 & 30 & 1920x1080 & 1,050 (00:35) & 80 & 47,557 & 45.3  & static & high & night & new \\
    MOT16-05 & 14 & 640x480 & 837 (01:00) & 124 & 6,818 & 8.1 & moving & medium & sunny & \cite{Ess:2008:CVPR} \\
    MOT16-09 & 30 & 1920x1080 & 525 (00:18) & 25 & 5,257 & 10.0  & static & low & indoor	& new \\
    MOT16-10 & 30 & 1920x1080 & 654 (00:22) & 54 & 12,318 & 18.8  & moving & medium & night	& new \\
    MOT16-11 & 30 & 1920x1080 & 900 (00:30) & 67 & 9,174 & 10.2 & moving & medium & indoor & new \\
    MOT16-13 & 25 & 1920x1080 & 750 (00:30) & 68 & 11,450 & 15.3  & moving & high & sunny & new \\
      \hline
      \multicolumn{3}{|c|}{\bf Total training} & {\bf 5,316 (03:35)} & {\bf 512} & {\bf 110,407} & {\bf 20.8} & & & &   \\
    \hline
    \multicolumn{11}{c}{\vspace{1em}} \\
    \hline
     \multicolumn{11}{|c|}{\bf Testing sequences} \\ 
     \hline 
 Name & FPS & Resolution & Length & Tracks & Boxes & Density &  Camera & Viewpoint & Conditions & Source\\ 
 \hline
       MOT16-01 & 30 & 1920x1080 & 450 (00:15) & 23 & 6,395 & 14.2  & static  & medium & cloudy & new \\
     MOT16-03 & 30 & 1920x1080 & 1,500 (00:50) & 148 & 104,556 & 69.7  & static & high & night & new \\
    MOT16-06 & 14 & 640x480 & 1,194 (01:25) & 217 & 11,538 & 9.7 & moving & medium & sunny & \cite{Ess:2008:CVPR} \\
    MOT16-07 & 30 & 1920x1080 & 500 (00:17) & 55 & 16,322 & 32.6  & moving & medium & shadow	& new \\
    MOT16-08 & 30 & 1920x1080 & 625 (00:21) & 63 & 16,737 & 26.8  & static & medium & sunny	& new \\
    MOT16-12 & 30 & 1920x1080 & 900 (00:30) & 94 & 8,295 & 9.2 & moving & medium & indoor & new \\
    MOT16-14 & 25 & 1920x1080 & 750 (00:30) &230 & 18,483 & 24.6  & moving & high & sunny & new \\
       \hline
      \multicolumn{3}{|c|}{\bf Total testing} & {\bf 5,919 (04:08)} & {\bf 830} & {\bf 182,326} & {\bf 30.8} & & & &  \\
      \hline 
    \end{tabular}
  \end{center}
    \caption{Overview of the sequences currently included in the \MOTNEW benchmark.}
\label{tab:dataoverview}
\end{table*}

\begin{table*}[tbh]
\begin {center}
 \begin{tabular}{|l| R{0.95cm}|R{0.95cm}|R{0.65cm}|R{0.85cm}|R{0.65cm}|R{0.85cm}|R{0.65cm}|R{0.65cm}|R{0.85cm}|R{0.85cm}|R{0.65cm}|R{0.99cm}|}
 \hline
 \multicolumn{13}{|c|}{\bf Annotation classes} \\ 
 \hline 
Sequence & Pedestrian & Person on vehicle   &  Car & Bicycle & Motorbike & Non motorized vehicle & Static person & 
Distractor  & Occluder on the ground & Occluder full      & Reflection   & Total \\
 \hline
  MOT16-01 & 6,395 & 346 & 0 & 341 & 0 & 0 & 4,790 & 900  &   3,150  & 0   &  0  & 15,922 \\ 
   MOT16-02  &  17,833 & 1,549  &  0   &  1,559 &  0  & 0  & 5,271  &  1,200 &  1,781  &  0  &  0  &    29,193\\
  MOT16-03  & 104,556  &  70  &  1,500  &   12,060  &  1,500     & 0   &  6,000   &  0     &  24,000    & 13,500    & 0   &  163,186\\
  MOT16-04  &  47,557   &    0   &  1,050   &  11,550  &   1,050  &   0  &  4,798  &   0   &   23,100  &  18,900  &  0  &    108,005\\
  MOT16-05 & 6,818 & 315   &  196 & 315 &  0   &  11  & 0  &  16  &   0    &   0 &  0  &   7,671\\
  MOT16-06 & 11,538  &   150  &  0     &  118  &     0    & 0   &   269  & 238 & 109     &  0  &   0      & 12,422\\
  MOT16-07  & 16,322   &   0   &  0  &  0  &  0    &  0  & 2,023   &  0  & 1,920     &    0    & 0    &  20,265\\
  MOT16-08 & 16,737  & 0  &  0   & 0 & 0   &  0  & 1,715 &  2,719 &   6,875  & 0  &  0  &     28,046\\
  MOT16-09  & 5,257  &  0  &  0     & 0  &  0 &  0 &  0   & 1,575 &  1,050  &   0   &  948     & 8,830\\
  MOT16-10  & 12,318  &  0  &  25  &  0    &  0 &  0  &   1,376 &  470   &   2,740   &  0  & 0    &   16,929\\
  MOT16-11  &  9,174   & 0  &  0 &  0   &  0  & 0  &  0   &  306  &   596  & 0  & 0      & 10,076\\
  MOT16-12  &   8,295  &  0  &  0   & 0   & 0    &  0 & 1,012  &  765 &    1,394    &  0     &  0   &   11,464\\
  MOT16-13   &  11,450   &    0  & 4,484   & 103   &  0    &  0    &  0   &  4    &   2,542  &    680     &    0   & 19,263\\
  MOT16-14  &  18,483    &   0   &  1,563    &     0  &    0 &   0   & 712  &   47   &  4,062     &     393   &  0   &     25,260\\
\hline
  Total & 292,733  & 2,430  & 8,818 & 26,046 &  2,550    & 11  &  27,966 & 8,238   & 73,319  & 33,473  & 948  &   476,532\\
      \hline 
    \end{tabular}
  \end{center}
    \caption{Overview of the types of annotations currently found in the \MOTNEW benchmark.}
\label{tab:dataclasses}
\end{table*}

\subsection{2D MOT 2016 sequences}

We have compiled a total of 14 sequences, of which we use half for 
training and half for testing. The annotations of the testing sequences 
will not be released in order to avoid (over)fitting of the methods to the 
specific sequences. 

Sequences are very different from each other, we can classify them according to:
\begin{itemize}
 \item Moving or static camera -- the camera can be held by a person, placed on a stroller or on a car, or can be positioned fixed in the scene.
 \item Viewpoint -- the camera can overlook the scene from a high position, a medium position (at pedestrian's height), or at a low position. 
 \item Conditions -- the weather conditions in which the sequence was taken are reported in order to obtain an estimate of the illumination conditions of the scene. Sunny sequences may contain shadows and saturated parts of the image, while the night sequence contains a lot of motion blur, making pedestrian detection and tracking rather challenging. Cloudy sequences on the other hand contain fewer of those artifacts.
\end{itemize}

              \begin{figure*}[tbh]
\centering
\subfigure[][ACF \cite{Dollar:2014:PAMI}]{
\includegraphics[width=4cm,height=3cm]{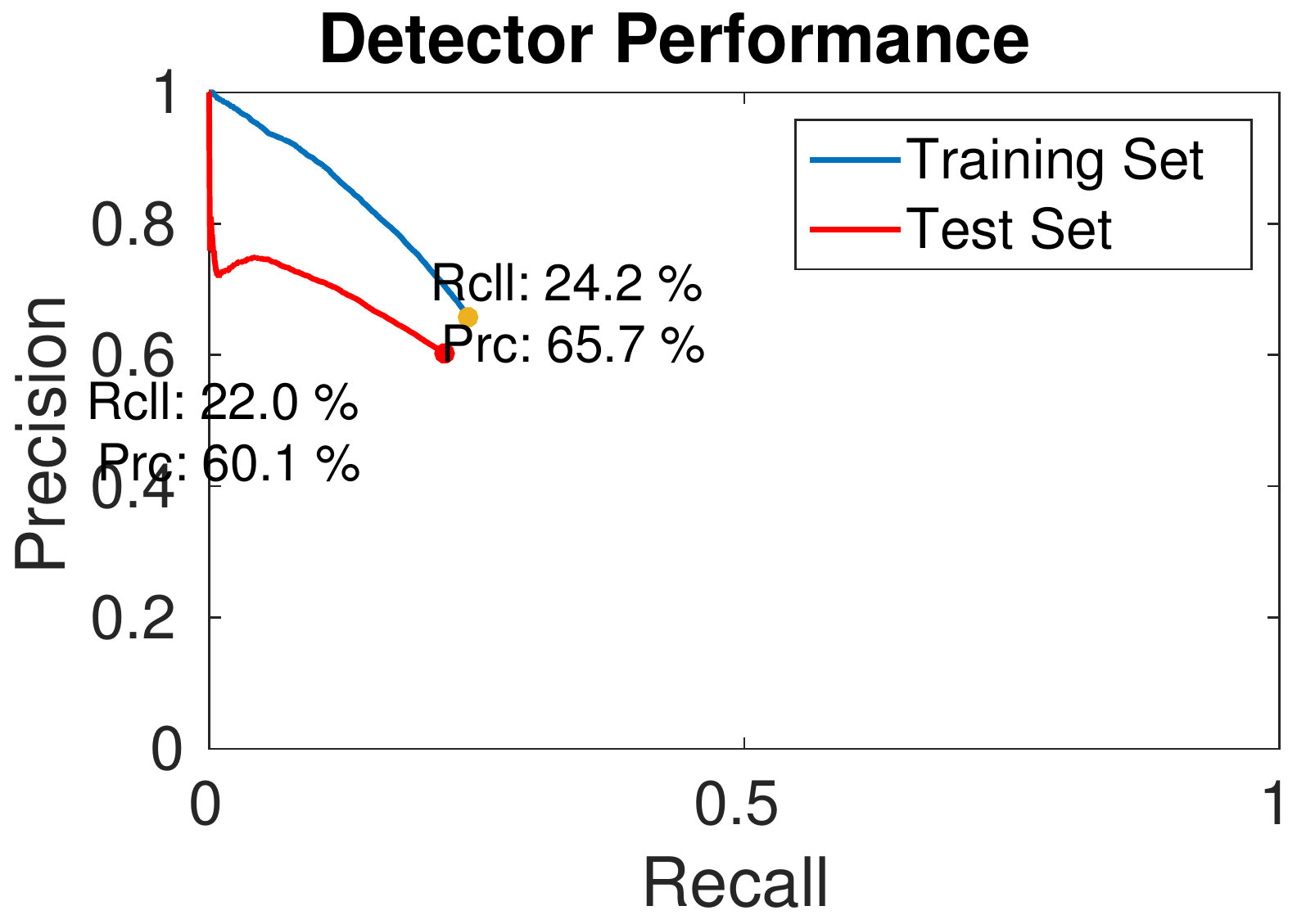}} 
\subfigure[][Fast-RCNN\cite{Girshick:2015:ICCV}]{
\includegraphics[width=4cm,height=3cm]{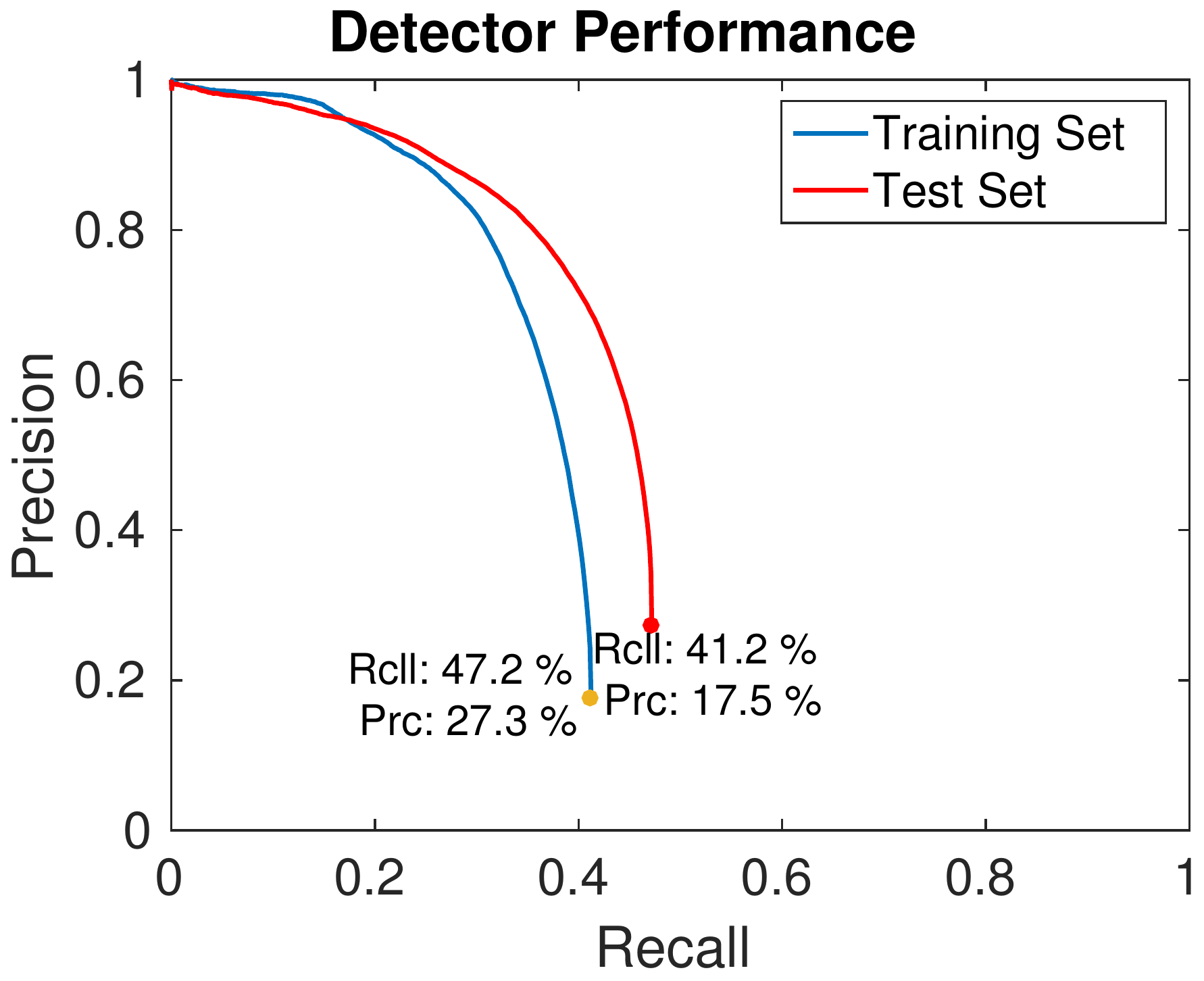}} 
\subfigure[][DPM v5\cite{Felzenszwalb:2010:PAMI}]{
\includegraphics[width=4cm,height=3cm]{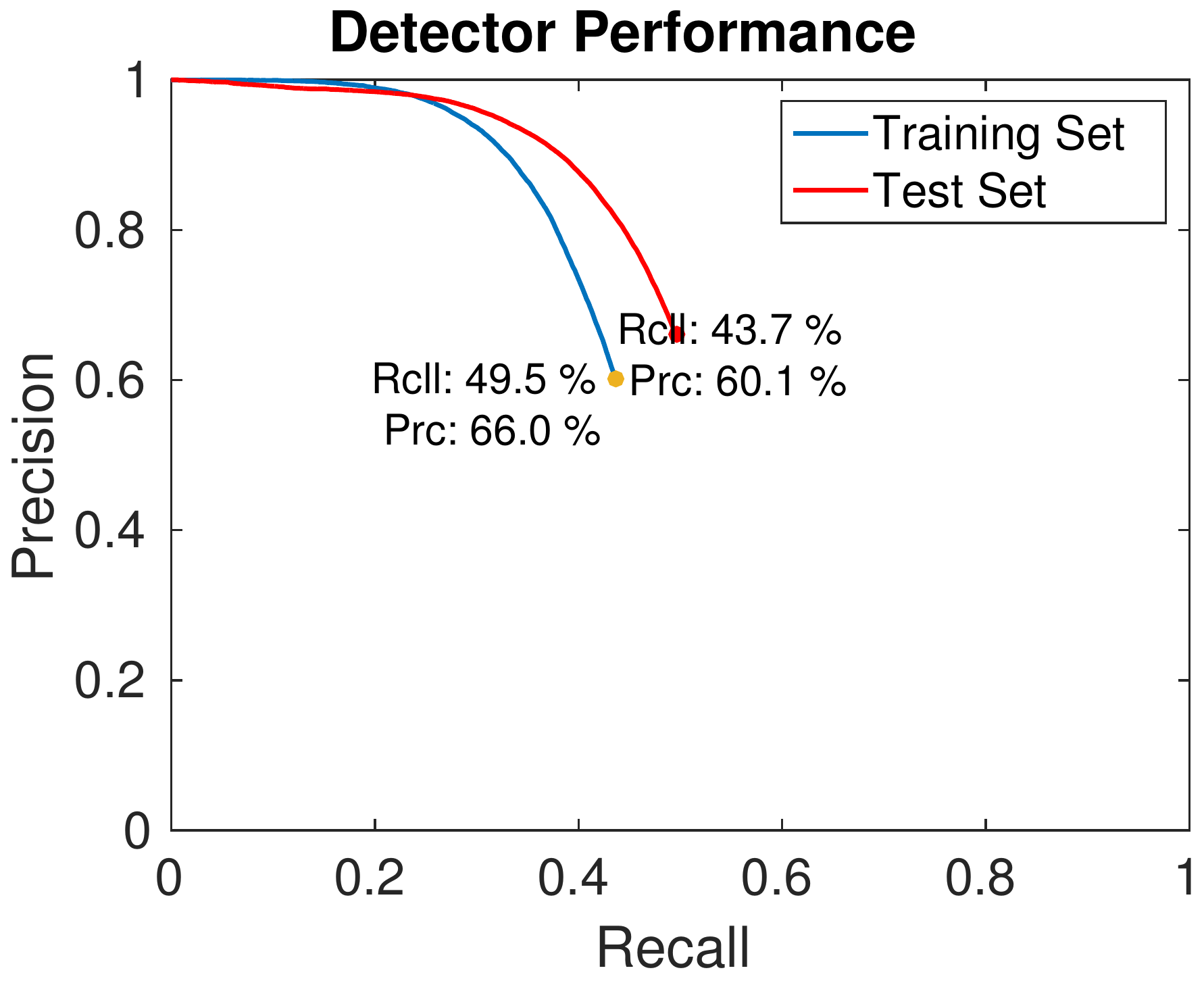}} 
\subfigure[][MOT16-03]{\raisebox{5mm}{
\includegraphics[width=4cm]{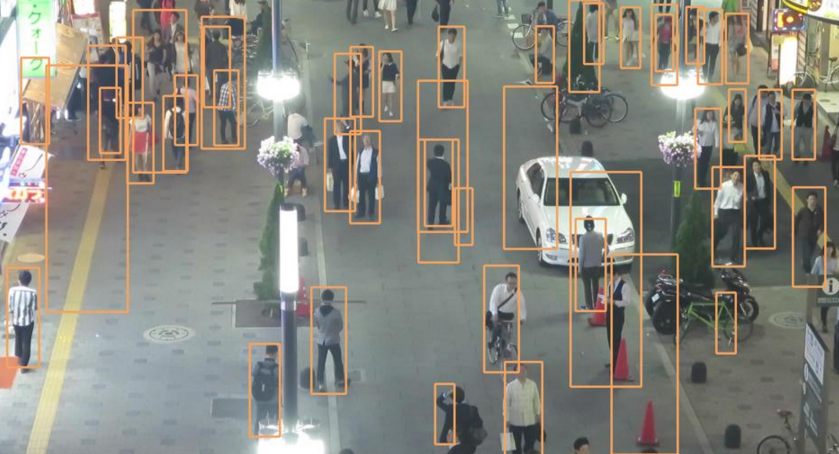}}} 
\caption{(a) The performance of three popular pedestrian detectors
evaluated on the training (blue) and the test (red) set. The circle indicates
the operating point (\ie the input detection set) for the trackers. (d) Exemplar detection results.
As can be seen, DPM provides the highest recall at the best precision, which is why we
 release only this detection set to the public.
}
\label{fig:det-performance}
\end{figure*}

The new data contains almost 3 times more bounding boxes for training and testing compared to \MOTOLD. Most sequences are filmed in high resolution, and the mean crowd density is 3 times higher when compared to the first benchmark release. Hence, we expect the new sequences to be more challenging for the tracking community.
In \Tab~\ref{tab:dataoverview}, we give an overview of the training and testing sequence characteristics for the challenge, including the number of bounding boxes used.

Aside from pedestrians, the annotations also include other classes like vehicles, bicycles, etc.~as detailed in \Sec~\ref{sec:anno-rules}. In \Tab~\ref{tab:dataclasses}, we detail the types of annotations that can be found in each sequence of \MOTNEW.

\subsection{Detections}
\label{sec:detections}

We tested several state-of-the-art detectors on our benchmark, obtaining the Precision-Recall curves in \Fig~\ref{fig:det-performance}. Note that the deformable part-based model (DPM) v5 \cite{Felzenszwalb:2010:PAMI, DPMv5} outperforms the other detectors in the task of pedestrian detection. 
As noted in \cite{girschickcvpr2015}, out-of-the-box R-CNN outperforms DPM in detecting all object classes except for the class ``person'', which is why we supply DPM detections with the benchmark.
We use the pretrained model with a low threshold
of $-1$ in order to maintain relatively high recall. 
Note that the recall does not reach 100\% 
because of the non-maximum suppression applied.
Exemplar detection results are shown in \Fig~\ref{fig:det-performance}.

A detailed breakdown of detection bounding boxes on individual sequences is provided in \Tab~\ref{tab:det-performance}.

\begin{table}[hbt]
\begin {center}
 \begin{tabular}{| l |r r r r |}
  \hline 
  \bf Seq & \bf nDet. & \bf nDet./fr. & \bf min height& \bf max height \\ 
          \hline 
  MOT16-01   &   3,775 &    8.39     &19.00    &258.92\\
  MOT16-02 &     7,267 &    12.11   &  19.00 &   341.97\\
  MOT16-03 &    85,854&     57.24  &   19.00&    297.57\\
  MOT16-04 &    39,437&     37.56  &   19.00&    341.97\\
  MOT16-05 &     4,333 &     5.18    & 19.00  &  225.27\\
  MOT16-06 &     7,851 &     6.58    & 19.00  &  210.12\\
  MOT16-07 &    11,309&     22.62   & 19.00 &   319.00\\
  MOT16-08 &    10,042&     16.07   &  19.00&    518.84\\
  MOT16-09 &     5,976 &    11.38    & 19.00 &   451.55\\
  MOT16-10 &     8,832 &    13.50    & 19.00 &   366.58\\
  MOT16-11 &     8,590 &     9.54     &19.00  &  518.84\\
  MOT16-12 &     7,764 &     8.63     &19.00  &  556.15\\
  MOT16-13 &     5,355 &     7.14     &19.00  &  210.12\\
  MOT16-14 &     8,781 &    11.71    & 19.00 &   258.92\\
\hline
      total  &  215,166   &  19.15  &   19.00   & 556.15\\      
 \hline 
    \end{tabular}
  \end{center}
    \caption{Detection bounding box statistics.}
\label{tab:det-performance}
\end{table}

Obviously, we cannot (nor necessarily want to) prevent anyone from
using a different set of detections, or relying on a different set
of features.
However, we require that this is noted as part of the tracker's 
description and is also displayed in the ratings table for transparency.

\subsection{Data format}
\label{sec:data-format}

All images were converted to JPEG and named sequentially to a 6-digit file name (\eg~000001.jpg). Detection and annotation files are simple comma-separated value (CSV) files. Each line represents one object instance and contains 9 values as shown in \Tab~\ref{tab:dataformat}.

The first number indicates in which frame the object appears, while
the second number identifies that object as belonging to a trajectory
by assigning a unique ID (set to $-1$ in a detection file, as no ID is
assigned yet). Each object can be assigned to only one trajectory.
The next four numbers indicate the position of the bounding box of the
pedestrian in 2D image coordinates. The position is indicated by the
top-left corner as well as width and height of the bounding box.
This is followed by a single number, which in case of detections
denotes their confidence score.
The last two numbers for detection files are ignored (set to -1).

\begin{table*}[hbt]
\begin {center}
 \begin{tabular}{| c | c| p{13cm}|}
 \hline
     \bf Position & \bf Name & \bf Description\\ 
          \hline 
    1 & Frame number & Indicate at which frame the object is present\\
 2 & Identity number & Each pedestrian trajectory is identified by a
 unique ID ($-1$ for detections)\\
 3 & Bounding box left &  Coordinate of the top-left corner of the pedestrian bounding box\\
  4 & Bounding box top & Coordinate of the top-left corner of the pedestrian bounding box \\
 5 & Bounding box width & Width in pixels of the pedestrian bounding box\\
 6 & Bounding box height & Height in pixels of the pedestrian bounding box\\
 7 & Confidence score & DET: Indicates how confident the detector is that this instance is a pedestrian. \hspace{5cm} GT: It acts as a flag whether the entry is to be considered (1) or ignored (0).  \\
  8 &  Class & GT: Indicates the type of object annotated  \\
 9 & Visibility & GT: Visibility ratio, a number between 0 and 1 that says how much of that object is visible. Can be due to occlusion and due to image border cropping. \\
      \hline 
    \end{tabular}
  \end{center}
    \caption{Data format for the input and output files, both for detection (DET) and annotation/ground truth (GT) files.}
\label{tab:dataformat}
\end{table*}

\begin{table}[hbt]
\begin {center}
 \begin{tabular}{| l | c|}
  \hline 
  \bf Label & \bf ID\\ 
          \hline 
Pedestrian & 1 \\
Person on vehicle & 2 \\
Car & 3 \\
Bicycle & 4 \\
Motorbike & 5 \\
Non motorized vehicle & 6 \\
Static person & 7 \\
Distractor & 8 \\
Occluder & 9 \\
Occluder on the ground & 10 \\
Occluder full & 11 \\
Reflection & 12 \\
 \hline 
    \end{tabular}
  \end{center}
    \caption{Label classes present in the annotation files and ID appearing in the 7$^\text{th}$ column of the files as described in \Tab~\ref{tab:dataformat}.}
\label{tab:labelclass}
\end{table}

 An example of such a 2D detection file is:
 \begin{samepage}
\begin{center}
\begin{footnotesize}
  \texttt{1, -1, 794.2, 47.5, 71.2, 174.8, 67.5, -1, -1}\nopagebreak\\
  \texttt{1, -1, 164.1, 19.6, 66.5, 163.2, 29.4, -1, -1}\nopagebreak\\
  \texttt{1, -1, 875.4, 39.9, 25.3, 145.0, 19.6, -1, -1}\nopagebreak\\
  \texttt{2, -1, 781.7, 25.1, 69.2, 170.2, 58.1, -1, -1}\nopagebreak\\
\end{footnotesize}
\end{center}
\end{samepage}

For the ground truth and results files, the 7$^\text{th}$ value (confidence score) acts as a flag whether the entry is to be considered. A value of 0 means that this particular instance is ignored in the evaluation, while a value of 1 is used to mark it as active. 
The 8$^\text{th}$ number indicates the type of object annotated, following the convention of \Tab~\ref{tab:labelclass}. The last number shows the visibility ratio of each bounding box. This can be due to occlusion by another static or moving object, or due to image border cropping.

 An example of such an annotation 2D file is:
 \begin{samepage}
\begin{center}
\begin{footnotesize}
  \texttt{1, 1, 794.2, 47.5, 71.2, 174.8,  1,  1, 0.8}\nopagebreak\\
  \texttt{1, 2, 164.1, 19.6, 66.5, 163.2,  1,  1, 0.5}\nopagebreak\\
  \texttt{2, 4, 781.7, 25.1, 69.2, 170.2, 0, 12, 1.}\nopagebreak\\
\end{footnotesize}
\end{center}
\end{samepage}

In this case, there are 2 pedestrians in the first frame of the sequence, with identity tags 1, 2. 
In the second frame, we can see a reflection (class 12), which is to be considered by the evaluation script and will neither count as a false negative, nor as a true positive, independent of whether it is correctly recovered or not.
Note that all values including the bounding box are 1-based, \ie the top left corner corresponds to $(1,1)$.

To obtain a valid result for the entire benchmark, a separate CSV file following the format described above must be created for each sequence and called \texttt{``Sequence-Name.txt''}. All files must be compressed into a single ZIP file that can then be uploaded to be evaluated.

\section{Evaluation}
\label{sec:evaluation}
Our framework is a platform for fair comparison of state-of-the-art
tracking methods. By providing authors with standardized ground truth
data, evaluation metrics and scripts, as well as a set of precomputed detections, all methods are compared under the exact same conditions, thereby isolating the performance of the tracker from everything else.
In the following paragraphs, we detail the set of evaluation metrics that we provide in our benchmark.

\subsection{Evaluation metrics}
\label{sec:evaluation-metrics}
In the past, a large number of metrics for quantitative evaluation of 
multiple target tracking have been proposed \cite{Smith:2005:CVPRW, 
Stiefelhagen:2006:CLE, Bernardin:2008:CLE, Schuhmacher:2008:ACM, 
Wu:2006:CVPR, Li:2009:CVPR}. Choosing ``the right'' one is largely 
application dependent and the quest for a unique, general evaluation metric
is still ongoing. On the one hand, it is desirable to summarize the performance 
into one single number to enable a direct comparison. On the other hand,
one might not want to lose information about the individual errors made
by the algorithms and provide several performance estimates, which 
precludes a clear ranking.

Following a recent trend \cite{Milan:2014:PAMI, Bae:2014:CVPR, 
Wen:2014:CVPR}, we employ two sets of measures that have established 
themselves in the literature: The \emph{CLEAR} metrics proposed by 
Stiefelhagen \etal \cite{Stiefelhagen:2006:CLE}, and a set of track 
quality measures introduced by Wu and Nevatia \cite{Wu:2006:CVPR}.
The evaluation scripts used in our benchmark are publicly 
available.\footnote{\url{http://motchallenge.net/devkit}}

\subsubsection{Tracker-to-target assignment}
\label{sec:tracker-assignment}
There are two common prerequisites for quantifying the performance of a 
tracker. One is to determine for each hypothesized output, whether it is a 
true positive (TP) that describes an actual (annotated) target, or 
whether the output is a false alarm (or false positive, FP). This 
decision is typically made by thresholding based on a defined distance 
(or dissimilarity) measure $\dismeas$ (see 
\Sec~\ref{sec:distance-measure}). A target that is missed by any 
hypothesis is a false negative (FN). A good result is expected to have 
as few FPs and FNs as possible. Next to the absolute numbers, we also 
show the false positive ratio measured by the number of false alarms per 
frame (FAF), sometimes also referred to as false positives per image 
(FPPI) in the object detection literature.

\begin{figure*}[t]
\centering
\def\svgwidth{1\linewidth}
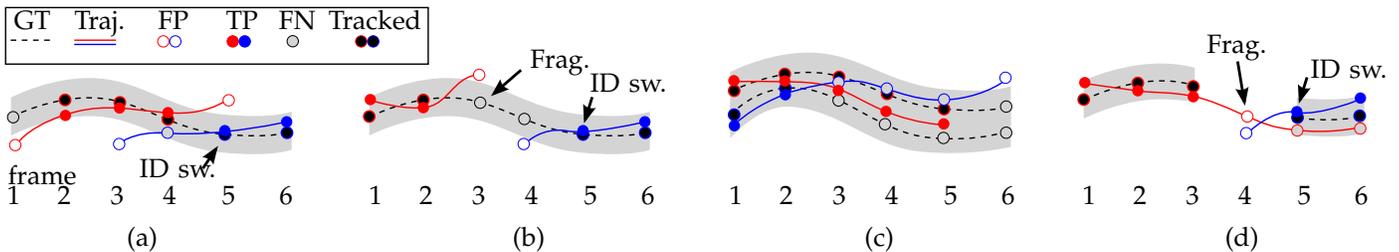
\caption
{
Four cases illustrating tracker-to-target assignments. (a) An ID switch 
occurs when the mapping switches from the previously assigned red track 
to the blue one. (b) A track fragmentation is counted in frame 3 because 
the target is tracked in frames 1-2, then interrupts, and then 
reacquires its `tracked' status at a later point. A new (blue) track hypothesis also 
causes an ID switch at this point. (c) Although the tracking results is 
reasonably good, an optimal single-frame assignment in frame 1 is 
propagated through the sequence, causing 5 missed targets (FN) and 4 
false positives (FP). Note that no fragmentations are counted in frames 
3 and 6 because tracking of those targets is not resumed at a later 
point. (d) A degenerate case illustrating that target re-identification 
is not handled correctly. An interrupted ground truth trajectory will 
typically cause a fragmentation. Also note the less intuitive ID switch, 
which is counted because blue is the closest target in frame 5 that is 
not in conflict with the mapping in frame 4. 
}
\label{fig:mapping}
\end{figure*}

Obviously, it may happen that the same target is covered by multiple 
outputs. The second prerequisite before computing the numbers is then to 
establish the correspondence between all annotated and hypothesized 
objects under the constraint that a true object should be recovered at 
most once, and that one hypothesis cannot account for more than one 
target. 

For the following, we assume that each ground truth trajectory has one 
unique start and one unique end point, \ie that it is not fragmented. 
Note that the current evaluation procedure does not explicitly handle 
target re-identification. In other words, when a target leaves the 
field-of-view and then reappears, it is treated as an unseen target with 
a new ID. As proposed in \cite{Stiefelhagen:2006:CLE}, the optimal 
matching is found using Munkre's (a.k.a.~Hungarian) algorithm. However, 
dealing with video data, this matching is not performed independently 
for each frame, but rather considering a temporal correspondence.
More precisely, if a ground truth object $i$ is matched to hypothesis 
$j$ at time $t-1$ \emph{and} the distance (or dissimilarity) between $i$ 
and $j$ in frame $t$ is below $\simthresh$, then the correspondence 
between $i$ and $j$ is carried over to frame $t$ even if there exists another
hypothesis that is closer to the actual target. A mismatch error (or 
equivalently an identity switch, IDSW) is counted if a ground truth 
target $i$ is matched to track $j$ and the last known assignment was $k 
\ne j$. Note that this definition of ID switches is more similar to 
\cite{Li:2009:CVPR} and stricter than the original one 
\cite{Stiefelhagen:2006:CLE}. Also note that, while it is certainly 
desirable to  keep the number of ID switches low, their absolute number 
alone is not always expressive to assess the overall performance, but 
should rather be considered in relation to the number of recovered 
targets. The intuition is that a method that finds twice as many 
trajectories will almost certainly produce more identity switches. For 
that reason, we also state the relative number of ID switches, which is 
computed as IDSW / Recall.

These relationships are illustrated in \Fig~\ref{fig:mapping}. For 
simplicity, we plot ground truth trajectories with dashed curves, and 
the tracker output with solid ones, where the color represents a unique 
target ID. The grey areas indicate the matching threshold (see next 
section). Each true target that has been successfully recovered in one 
particular frame is represented with a filled black dot with a stroke 
color corresponding to its matched hypothesis. False positives and false 
negatives are plotted as empty circles. See figure caption for more 
details.

After determining true matches and establishing correspondences it
is now possible to compute the metrics. We do so by concatenating all
test sequences and evaluating on the entire benchmark. This is in
general more meaningful instead of averaging per-sequences figures due to
the large variation in the number of targets.

\subsubsection{Distance measure}
\label{sec:distance-measure}

In the most general case, the relationship between ground truth objects 
and a tracker output is established using bounding boxes on the image 
plane. Similar to object detection \cite{Everingham:2012:VOC}, the 
intersection over union (a.k.a. the Jaccard index) is usually employed 
as the similarity criterion, while the threshold $\simthresh$ is set to 
$0.5$ or $50\%$.

\subsubsection{Target-like annotations}

\begin{figure*}
\centering
 \includegraphics[height=3.6cm]{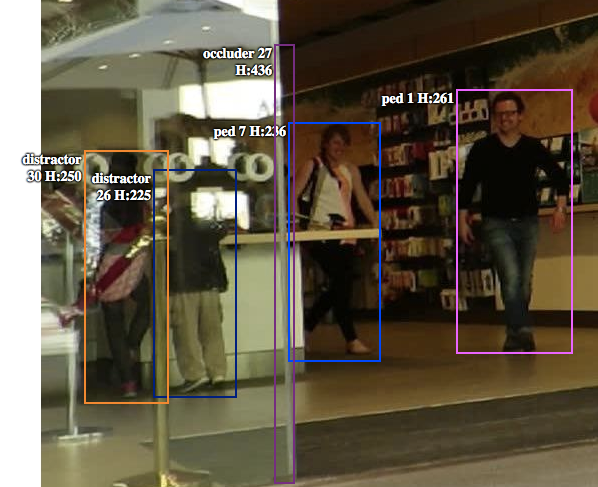}
  \includegraphics[height=3.6cm]{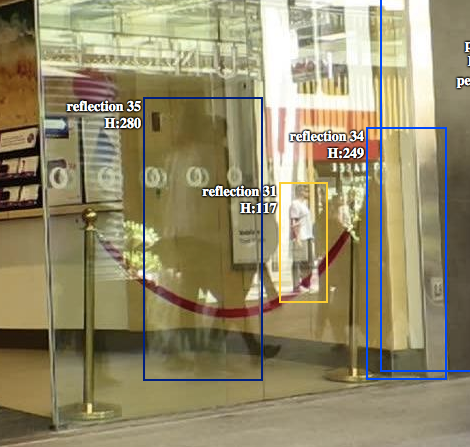}
   \includegraphics[height=3.6cm]{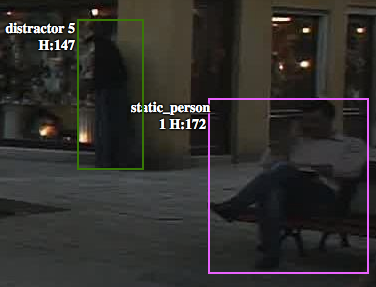}
  \includegraphics[height=3.6cm]{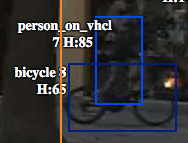}
 \caption{The annotations include different classes of objects similar to the target class, a pedestrian in our case. We consider these special classes (distractor, reflection, static person and person on vehicle) to be so similar to the target class that a tracker should neither be penalized nor rewarded for tracking them in the sequence.}
 \label{fig:distractors}
\end{figure*}

People are a common object class present in many scenes, but should we track all people in our benchmark?
For example, should we track static people sitting on a bench? Or people on bicycles? How about people behind a glass?  
We define the target class of \MOTNEW as all upright people, standing or walking, that are reachable along the viewing ray without a physical obstacle, \ie reflections, people behind a transparent wall or window are excluded.
We also exclude from our target class people on bycicles or other vehicles.
For all these cases where the class is very similar to our target class (see Figure \ref{fig:distractors}), we adopt a similar strategy as in \cite{Mathias:2014:ECCV}. That is, a method is neither penalized nor rewarded for tracking or not tracking those similar classes. 
Since a detector is likely to fire in those cases, we do not want to penalize a tracker with a set of false positives for properly following that set of detections, \ie of a person on a bicycle. Likewise, we do not want to penalize with false negatives a tracker that is based on motion cues and therefore does not track a sitting person.

In order to handle these special cases, we adapt the tracker-to-target assignment algorithm to perform the following steps:

\begin{enumerate}
\item At each frame, all bounding boxes of the result file are matched to the ground truth via the Hungarian algorithm.
\item All result boxes that overlap $>50\%$ with one of these classes (distractor, static person, reflection, person on vehicle) are removed from the solution.
\item During the final evaluation, {\it only} those boxes that are annotated as {\it pedestrians} are used.
\end{enumerate}

\subsubsection{Multiple Object Tracking Accuracy}
\label{sec:mota}
The MOTA \cite{Stiefelhagen:2006:CLE} is perhaps the most widely used 
metric to evaluate a tracker's performance. The main reason for this is 
its expressiveness as it combines three sources of errors defined above:
\begin{equation}
\text{MOTA} = 
1 - \frac
{\sum_t{(\text{FN}_t + \text{FP}_t + \text{IDSW}_t})}
{\sum_t{\text{GT}_t}},
\label{eq:mota}
\end{equation}
where $t$ is the frame index and GT is the number of ground truth 
objects. We report the percentage MOTA $(-\infty, 100]$ in our 
benchmark. Note that MOTA can also be negative in cases where the number 
of errors made by the tracker exceeds the number of all objects in the 
scene.

Even though the MOTA score gives a good indication of the overall 
performance, it is highly debatable whether this number alone can serve 
as a single performance measure. 

{\bf{Robustness.}}
One incentive behind compiling this benchmark was to reduce dataset bias
by keeping the data as diverse as possible. The main motivation is to
challenge state-of-the-art approaches and analyze their performance in
unconstrained environments and on unseen data. Our experience shows that
most methods can be heavily overfitted on one particular dataset, and may not be general enough to handle an entirely
different setting without a major change in parameters or even in the
model.

To indicate the robustness of each tracker across \emph{all} benchmark
sequences, we show the standard deviation of their MOTA score.

\subsubsection{Multiple Object Tracking Precision}
\label{sec:motp}

The Multiple Object Tracking Precision is the average dissimilarity 
between all true positives and their corresponding ground truth targets. 
For bounding box overlap, this is computed as 
\begin{equation}
\text{MOTP} = 
\frac
{\sum_{t,i}{d_{t,i}}}
{\sum_t{c_t}},
\label{eq:motp}
\end{equation}
where $c_t$ denotes the number of matches in frame $t$ and $d_{t,i}$ is 
the bounding box overlap of target $i$ with its assigned ground truth 
object. MOTP thereby gives the average overlap between all correctly 
matched hypotheses and their respective objects and ranges between 
$\simthresh := 50\%$ and $100\%$.

It is important to point out that MOTP is a 
measure of localization precision, \emph{not} to be confused with the 
\emph{positive predictive value} or \emph{relevance} in the context of 
precision / recall curves used, \eg, in object detection.

In practice, it mostly quantifies the localization accuracy of the detector, 
and therefore, it provides little information about the actual performance of the tracker.


\subsubsection{Track quality measures}
\label{sec:track-measures}


Each ground truth trajectory can be classified as mostly tracked (MT), 
partially tracked (PT), and mostly lost (ML). This is done based on how 
much of the trajectory is recovered by the tracking algorithm. A target 
is mostly tracked if it is successfully tracked for at least $80\%$ of 
its life span. Note that it is irrelevant for this measure whether the 
ID remains the same throughout the track. If a track is only recovered 
for less than $20\%$ of its total length, it is said to be mostly lost 
(ML). All other tracks are partially tracked. A higher number of MT and 
few ML is desirable. We report MT and ML as a ratio of mostly tracked 
and mostly lost targets to the total number of ground truth 
trajectories.

In certain situations one might be interested in obtaining long, 
persistent tracks without gaps of untracked periods. To that end, the 
number of track fragmentations (FM) counts how many times a ground truth 
trajectory is interrupted (untracked). In other words, a fragmentation 
is counted each time a trajectory changes its status from tracked to 
untracked and tracking of that same trajectory is resumed at a later 
point. Similarly to the ID switch ratio 
(\cf~\Sec~\ref{sec:tracker-assignment}), we also provide the relative 
number of fragmentations as FM / Recall.

\subsubsection{Tracker ranking}
\label{sec:ranking}
As we have seen in this section, there are a number of reasonable 
performance measures to assess the quality of a tracking system, which 
makes it rather difficult to reduce the evaluation to one single number. 
To nevertheless give an intuition on how each tracker performs compared 
to its competitors, we compute and show the average rank for each one by 
ranking all trackers according to each metric and then averaging across 
all performance measures.
%

\section{Baseline Methods}
\label{sec:baselines}
As a starting point for the benchmark, we are working on including a number of 
recent multi-target tracking approaches as baselines, which we will 
briefly outline for completeness but refer the reader to the respective 
publication for more details. Note that we have used the publicly 
available code and trained all of them in the same way
(\cf~\Sec~\ref{sec:training}). 
However, we explicitly state that the 
provided numbers may not represent the best possible performance for 
each method, as could be achieved by the authors themselves. 
Table~\ref{tab:results} lists current benchmark results for all 
baselines as well as for all anonymous entries at the time of 
writing of this manuscript.

\begin{table*}[tb]
\begin {center}
\setlength{\tabcolsep}{0.15cm}
\begin{tabular}{l| rrrrrrrrrrrrr}
Method &  MOTA & MOTP & FAR & MT(\%) & ML(\%) & FP & FN & IDsw & rel.ID & FM & rel.FM & Hz & Ref.\\
\hline
\Tracker{TBD} & 33.7 {\tiny $\pm$9.2} & 76.5 & 1.0 & 7.2 & 54.2 & 5,804 & 112,587 & 2,418 & 63.3 & 2,252 & 58.9 & 1.3 & \cite{Geiger:2014:PAMI}\\
\Tracker{CEM} & 33.2 {\tiny $\pm$7.9} & 75.8 & 1.2 & 7.8 & 54.4 & 6,837 & 114,322 & 642 & 17.2 & 731 & 19.6 &  0.3 & \cite{Milan:2014:PAMI}\\
\Tracker{DP\_NMS} & 32.2 {\tiny $\pm$9.8} & 76.4 & 0.2 & 5.4 & 62.1 & 1,123 & 121,579 & 972 & 29.2 & 944 & 28.3 & 212.6 & \cite{Pirsiavash:2011:CVPR} \\
\Tracker{SMOT} & 29.7 {\tiny $\pm$7.3} & 75.2 & 2.9 & 4.3 & 47.7 & 17,426 & 107,552 & 3,108 & 75.8 & 4,483 & 109.3 & 0.2 &  \cite{Dicle:2013:ICCV}\\
\Tracker{JPDA\_m} & 26.2 {\tiny $\pm$6.1} & 76.3 & 0.6 & 4.1 & 67.5 & 3,689 & 130,549 & 365 & 12.9 & 638 & 22.5 & 22.2 & \cite{Rezatofighi:2015:ICCV} \\

\end{tabular}
\caption{Quantitative results of the baselines on MOT16.}
\label{tab:results}
\end{center}
\end{table*}

\subsection{Training and testing}
\label{sec:training}
Most of the available tracking approaches do not include a learning (or 
training) algorithm to determine the set of model parameters for a 
particular dataset. Therefore, we follow a simplistic search scheme for 
all baseline methods to find a good setting for our benchmark. To that 
end, we take the default parameter set $\parvec := 
\{\onepar_1,\ldots,\onepar_P\}$ as suggested by the authors, where $P$ is 
the number of free parameters for each method. We then perform 20 
independent runs 
on the training set with varying parameters. In each run, a parameter 
value $\onepar_i$ is uniformly sampled around its default value in the 
range $[\frac{1}{2}\onepar_i , 2 \onepar_i]$. Finally, the parameter set 
$\parvec^*$ that achieved the highest MOTA score across all 20 runs
(\cf~\Sec~\ref{sec:mota}) is taken as the optimal setting and run once 
on the test set. The optimal parameter set is stated in the description
entry for each baseline method on the benchmark website.

\subsection{\Tracker{DP\_NMS}: Network flow tracking}
\label{sec:DP_NMS}
Since its original publication \cite{Zhang:2008:CVPR}, a large number of 
methods that are based on the network flow formulation have appeared in 
the literature \cite{Pirsiavash:2011:CVPR, Butt:2013:CVPR, 
Liu:2013:CVPR, Wang:2014:CVPR, lealcvpr2012}. The basic idea is to model the tracking 
as a graph, where each node represents a detection and each edge 
represents a transition between two detections. Special source and sink 
nodes allow spawning and absorbing trajectories. A solution is obtained 
by finding the minimum cost flow in the graph. Multiple assignments and track 
splitting is prevented by introducing binary and linear constraints.

Here we use two solvers: (i) the successive shortest paths approach 
\cite{Pirsiavash:2011:CVPR} that employs dynamic programming with
non-maxima suppression, termed \Tracker{DP\_NMS}; (ii) a linear programming solver 
that appears as a baseline in \cite{lealcvpr2014}.
This solver uses the Gurobi Library \cite{gurobi}.

\subsection{\Tracker{CEM}: Continuous energy minimization}
\label{sec:CEM}
\Tracker{CEM} \cite{Milan:2014:PAMI} formulates the problem in terms of a 
high-dimensional continuous energy. Here, we use the basic approach 
\cite{Andriyenko:2011:CVPR} without explicit occlusion reasoning or 
appearance model. The target state $\mathbf{X}$ is represented by 
continuous $x,y$ coordinates in \emph{all} frames. The energy 
$E(\mathbf{X})$ is made up of several components, including a data term 
to keep the solution close to the observed data (detections), a dynamic 
model to smooth the trajectories, an exclusion term to avoid collisions, 
a persistence term to reduce track fragmentations, and a regularizer.
The resulting energy is highly non-convex and is minimized in an alternating
fashion using conjugate gradient descent and deterministic jump moves.

\subsection{\Tracker{SMOT}: Similar moving objects}
\label{sec:SMOT}
The Similar Multi-Object Tracking (\Tracker{SMOT}) approach 
\cite{Dicle:2013:ICCV} specifically targets situations where target 
appearance is ambiguous and rather concentrates on using the motion as a 
primary cue for data association. Tracklets with similar motion are linked
to longer trajectories using the generalized linear assignment (GLA) 
formulation. The motion similarity and the underlying dynamics of a 
tracklet are modeled as the order of a linear regressor approximating that
tracklet.

\subsection{\Tracker{TBD}: Tracking-by-detection}
\label{sec:TBD}
This two-stage tracking-by-detection (\Tracker{TBD}) approach 
\cite{Geiger:2014:PAMI, Zhang:2013:ICCV} is part of a larger traffic 
scene understanding framework and employs a rather simple data 
association technique. The first stage links overlapping detections
with similar appearance in successive frames into tracklets. The
second stage aims to bridge occlusions of up to 20 frames. Both stages
employ the Hungarian algorithm to optimally solve the matching problem.
Note that we did not re-train this baseline but rather used the original
implementation and parameters provided.

%
%
%

\subsection{\Tracker{JPDA\_m}: Joint probabilistic data association using $m$-best solutions}
\label{sec:JPDAm}
Joint Probabilistic Data Association (JPDA)~\cite{Fortman:1980:MTT} is one of the oldest 
techniques for global data association. It first builds a joint hypothesis
probability that includes \emph{all} possible assignments, and then computes the
marginal probabilities for each target to be assigned to each measurement.
The state for each target is estimated in an online manner, typically using
one Kalman filter per target. While this approach is theoretically sound, it
is prohibitive in practice due to the exponential number of possible assignment
hypotheses to be considered. It has thus long been considered impractical for
computer vision applications, especially in crowded scenes.
Recently, an efficient approximation to JPDA was proposed~\cite{Rezatofighi:2015:ICCV}.
The main idea is to approximate the full joint probability distribution by $m$ strongest
hypotheses. It turns out that in practice, only around 100 most likely assignments
are sufficient to obtain the exact same solution as full JPDA.

 \section{Conclusion and Future Work}
 \label{sec:conclusion}
 
We have presented a new challenging set of sequences within the \MOTChallenge benchmark.
The 2016 sequences contain 3 times more targets to be tracked when compared to the initial 2015 version. 
Furthermore, more accurate annotations were carried out following a strict protocol, and extra classes such as vehicles, 
sitting people, reflections or distractors were also annotated to provide further information to the community.
We believe that the \MOTNEW release within the already established \MOTChallenge benchmark
provides a fairer comparison of state-of-the-art tracking methods, and challenges researchers to develop more generic
methods that perform well in unconstrained
environments and on unseen data.
In the future, we plan to continue our workshops and challenges series, 
and also introduce various other \mbox{(sub-)}benchmarks for targeted
applications, \eg sport analysis, or biomedical cell tracking. \\

\noindent{\bf Acknowledgements.} We would like to specially acknowledge 
Siyu Tang, Sarah Becker, Andreas Lin and Kinga Milan for their help in 
the annotation process. We thank Bernt Schiele for helpful discussions
and important insights into benchmarking. 
IDR gratefully acknowledges the support of the 
Australian Research Council through FL130100102.

\ifCLASSOPTIONcaptionsoff
  \newpage
\fi

\bibliographystyle{ieee}

\bibliography{refs-lau,refs-short,refs-anton,refs-new}


\end{document}